\documentclass[conference,a4paper]{ieeeconf}

\IEEEoverridecommandlockouts
\usepackage[left=45pt,right=45pt,top=54pt,bottom=72pt]{geometry}
\usepackage{graphicx}
\usepackage{amsmath}
\usepackage{amssymb}
\usepackage{amsfonts}
\usepackage{array}
\usepackage{booktabs}
\usepackage{multirow}
\usepackage{url}
\usepackage{cite}
\usepackage{float}
\usepackage{subfigure}
\usepackage{textcomp}

\title{\LARGE \bf
A Rapid Deployment Pipeline for Autonomous Humanoid Grasping  Based on Foundation Models
}

\author{
\authorblockN{
Yifei Yan$^{1*}$, 
Yankai Liao$^{2*}$, 
and Linqi Ye$^{1}$
}
\thanks{*These authors contributed equally to this work.}
\thanks{$^{1}$Y. Yan and L. Ye are with the School of Future Technology, Shanghai University, Shanghai, 200444, China (Corresponding author, email: yelinqi@shu.edu.cn)
}
\thanks{$^{2}$Y. Liao is with the School of Mechanical Engineering, Shanghai Institute of Technology, Shanghai, 201418, China}
}

\begin{document}

\maketitle
\thispagestyle{empty}
\pagestyle{empty}

\begin{abstract}
Deploying a humanoid robot to manipulate a new object has traditionally required one to two days of effort: data collection, manual annotation, 3D model acquisition, and model training. This paper presents an end-to-end rapid deployment pipeline that integrates three foundation-model components to shorten the onboarding cycle for a new object to approximately 30 minutes: (i) Roboflow-based automatic annotation to assist in training a YOLOv8 object detector; (ii) 3D reconstruction based on Meta SAM 3D, which eliminates the need for a dedicated laser scanner; and (iii) zero-shot 6-DoF pose tracking based on FoundationPose, using the SAM~3D-generated mesh directly as the template. The estimated pose drives a Unity-based inverse kinematics planner, whose joint commands are streamed via UDP to a Unitree~G1 humanoid and executed through the Unitree SDK. We demonstrate detection accuracy of mAP@0.5 = 0.995, pose tracking precision of $\sigma < 1.05$ mm, and successful grasping on a real robot at five positions within the workspace. We further verify the generality of the pipeline on an automobile-window glue-application task. The results show that combining foundation models for perception with everyday imaging devices (e.g., smartphones) can substantially lower the deployment barrier for humanoid manipulation tasks.
\end{abstract}


\section{INTRODUCTION}

Humanoid robots are increasingly being deployed in human-centered environments, where they must manipulate a wide variety of everyday objects---bottles, tools, door handles, assembly-line parts. Enabling robots to quickly adapt to new objects remains a core bottleneck. The traditional deployment workflow for a new object consists of four stages: (a) collecting a visual dataset, (b) manually annotating bounding boxes, (c) acquiring an accurate 3D mesh---typically requiring professional structured-light or laser scanners, and (d) training perception and grasp-planning models. In practice, this pipeline typically consumes several days per object and relies on specialized hardware that is inconvenient for on-site deployment or rapid iteration.

Recent advances in foundation models offer an opportunity to compress this timeline. Vision foundation models such as the Segment Anything Model (SAM)~\cite{kirillov2023segment} enable automatic annotation tools like Roboflow to produce bounding-box labels with minimal human intervention. Zero-shot pose estimators such as FoundationPose~\cite{wen2024foundationpose} can track arbitrary objects at 30 fps given only an RGB-D stream and a 3D mesh, with no per-object retraining. More recently, 3D reconstruction systems such as Meta SAM~3D~\cite{chen2025sam3d} can generate textured meshes from a single photograph, completely removing the dependence on specialized scanners. In this paper, we integrate these components into a closed-loop humanoid manipulation system and quantitatively evaluate the resulting improvement in deployment efficiency and end-to-end reliability.

We propose a unified pipeline that chains together Roboflow-assisted YOLOv8 training, SAM~3D reconstruction, and FoundationPose tracking to form the real-time perception system for a Unitree~G1 humanoid robot. The estimated pose is consumed by a Unity-based inverse-kinematics planner, whose joint commands are streamed over UDP to a C++ bridge running the Unitree SDK. We further validate the generality of the pipeline on an automobile-window glue-application task, in which the same architecture is reused simply by replacing the object mesh and the YOLO class label.

Compared with the currently prevailing Vision-Language-Action (VLA) \cite{zitkovich2023rt2,kim2024openvla} end-to-end paradigm, the framework proposed in this paper adopts a modular design, requires no training data, and offers stronger interpretability and debuggability, making it more suitable for precision manipulation deployment of humanoid robots in real-world scenarios. Existing VLA models (e.g., RT-2, OpenVLA) generally rely on massive pretraining data and model fine-tuning, suffer from high inference latency and high deployment barriers on real robots, and are difficult to apply directly to industrial and service scenarios that demand high stability.

The contributions of this paper can be summarized as follows:
\begin{enumerate}
\item We integrate SAM~3D reconstruction with FoundationPose-based 6-DoF tracking for humanoid manipulation, replacing the laser-scanning pipeline with a camera-based capture setup.
\item We validate the reusability of the pipeline by applying the same architecture to a structurally different second task (window glue application).
\end{enumerate}

\section{RELATED WORK}

\subsection{Zero-Shot 6-DoF Pose Estimation}

Deep-learning-based 6-DoF pose estimators, such as PoseCNN~\cite{xiang2017posecnn} and PVNet~\cite{peng2022pvnet}, require per-object training and depend on annotated pose datasets. Foundation-model approaches have recently begun to remove this requirement. MegaPose~\cite{labbe2022megapose} performs render-and-compare matching for novel objects given a CAD mesh. FoundationPose further combines neural implicit representations~\cite{mildenhall2021nerf} with a transformer-based refinement network, achieving zero-shot tracking on RGB-D input at 30 fps; this is the perception backbone adopted in our pipeline. BundleSDF~\cite{wen2023bundlesdf} supports the complementary model-free setting by reconstructing object meshes online from reference views, but at substantially higher computational cost.

\subsection{Foundation Models for Detection and 3D Reconstruction}

SAM, as a promptable segmentation network, generalizes well across many domains; its core role is to produce pixel-level masks of the target. Automatic annotation tools such as Roboflow can compute bounding boxes from these masks, thereby generating detection labels and dramatically reducing manual annotation cost. YOLOv8~\cite{yaseen2024yolov8} adopts a decoupled detection head and C2f modules, achieving a good balance between accuracy and speed in lightweight models. For 3D reconstruction, the recently released SAM~3D can generate textured meshes from a single monocular image; we adopt this approach in place of structured-light scanning, effectively lowering the barrier for 3D model acquisition.

\subsection{Humanoid Manipulation and Sim-to-Real Transfer}

Recent humanoid platforms, such as the Unitree~G1, have made research on bimanual manipulation more accessible. Ze~et~al.~\cite{ze2025generalizable} demonstrated 3D diffusion policies for humanoid tasks, and Open-TeleVision~\cite{cheng2024opentelevision} leverages teleoperation for data collection. In contrast, our work aims at a training-free perception-to-action path by combining foundation models with classical inverse kinematics, borrowing ideas from domain randomization~\cite{tobin2017domain} only in low-level motion smoothing.

\section{METHOD}

\begin{figure}[htbp]
\centering
\includegraphics[width=0.95\columnwidth]{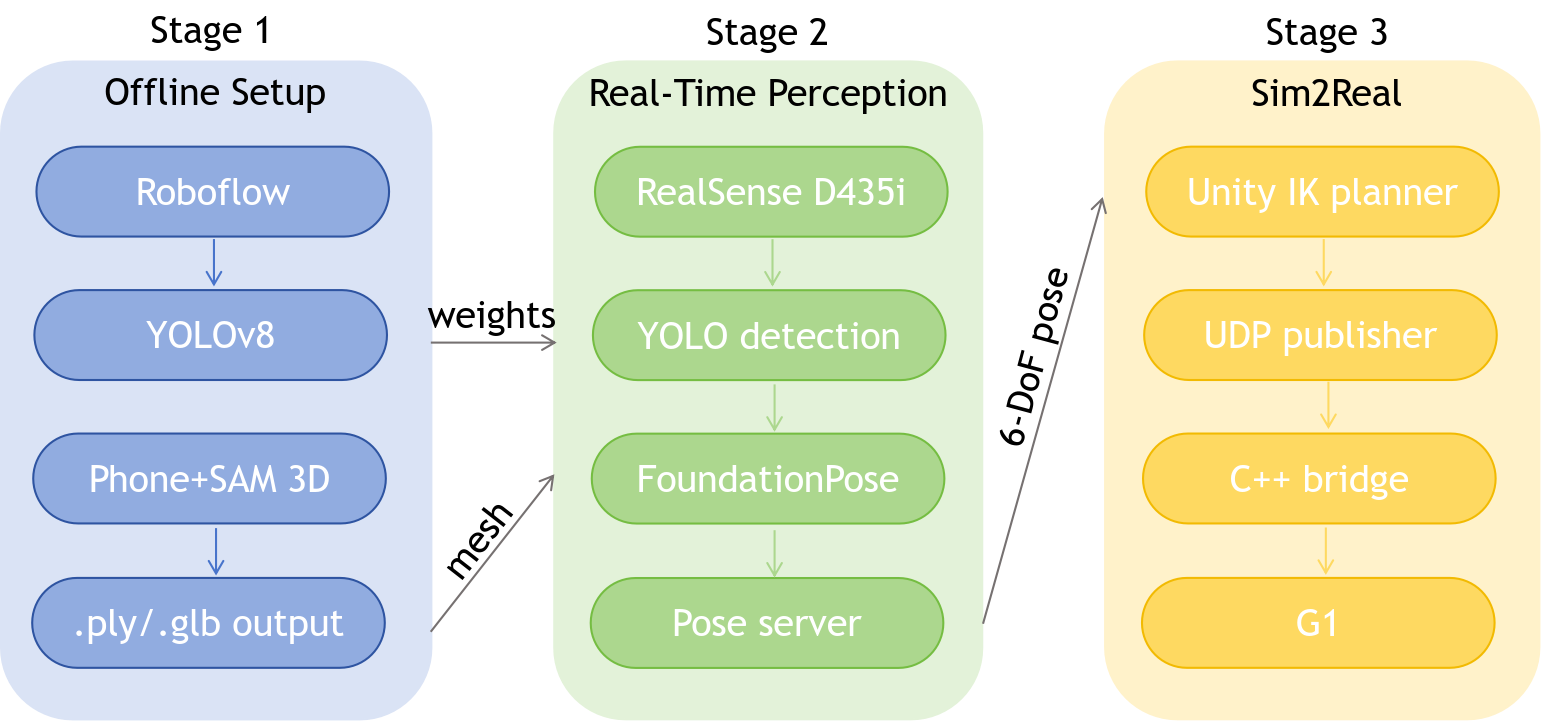}
\caption{The three-stage pipeline for rapid deployment of humanoid grasping.}
\label{fig:pipeline}
\end{figure}

Our pipeline consists of three stages, chained together in a producer--consumer fashion: an offline asset-preparation stage (Sec.~\ref{sec:offline}), an online perception stage running at the camera frame rate (Sec.~\ref{sec:online}), and a sim-to-real execution stage that consumes the perception output (Sec.~\ref{sec:sim2real}). Fig.~\ref{fig:pipeline} shows the overall architecture.

\subsection{Offline Asset Preparation}
\label{sec:offline}

\textbf{Automatic dataset annotation.} A RealSense~D435i~\cite{keselman2017realsense} camera is used to collect 150 images of the target object from diverse viewpoints. The images are uploaded to Roboflow, whose SAM-assisted auto-labeling feature generates bounding boxes. A human reviewer spot-checks and corrects failure cases. The final split is 105/30/15 images for training, validation, and testing, respectively.

\textbf{YOLOv8 training.} We fine-tune a COCO-pretrained YOLOv8n model with the following hyperparameters: 50 epochs, batch size 8, image size $640{\times}640$, trained on CPU with automatic mixed precision disabled, using the default data-augmentation policy.

\textbf{3D reconstruction.} A single photograph of the target object is uploaded to Meta SAM~3D to obtain a textured mesh. The mesh is exported in \texttt{.glb} format for Unity rendering and converted to \texttt{.obj} via Poisson surface reconstruction from a \texttt{.ply} file for FoundationPose. The entire reconstruction takes about 5 minutes per object.

\subsection{Online Perception}
\label{sec:online}

Each RGB-D frame is processed by a cascade of networks. The YOLOv8 detector is invoked once every 5 frames to reduce GPU load, producing a bounding box and confidence score for the target object. The image region covered by the bounding box is used to generate the 2D prior required by FoundationPose: specifically, for each bounding box we construct a rectangular region of interest (ROI), from which depth values and pixel coordinates are extracted as input to FoundationPose's pose-registration stage. FoundationPose operates in two modes:

\begin{enumerate}
\item \emph{Registration mode.} Three refinement iterations are performed at the first frame or after re-initialization.
\item \emph{Tracking mode.} One iteration is performed per frame, using the previous frame's pose as initialization.
\end{enumerate}

Each object tracker is managed by a three-state finite-state machine (\textsc{Uninitialized} $\rightarrow$ \textsc{Tracking} $\rightarrow$ \textsc{Lost}). The conditions for entering the \textsc{Lost} state are: (i) three consecutive tracking failures; (ii) a position change exceeding 0.15 m or a rotation change exceeding $90^\circ$ between two consecutive frames; (iii) no YOLO detection for more than 3 seconds. Recovery from \textsc{Lost} is triggered by a new YOLO detection, which re-supplies a bounding box to re-initialize tracking. The estimated pose is serialized to JSON and exposed on a local HTTP endpoint at 30 Hz.

\subsection{Sim-to-Real Execution}
\label{sec:sim2real}

A Unity scene containing the G1 URDF polls the perception endpoint at 30 Hz and mirrors the tracked object's pose onto a virtual twin. The inverse-kinematics controller plans a multi-stage grasping trajectory (pre-grasp lift, approach, descent, gripper close, lift, release), interpolating between waypoints with a smoothstep. Target joint angles are normalized to the $[0, 1]$ interval based on each joint's configuration range and published over UDP to a C++ bridge program.

The bridge recovers radian values via $\theta = \theta_{\min} + \bar{\theta}(\theta_{\max} - \theta_{\min})$, clamps them within joint limits, and issues commands via three channels of the Unitree SDK: \texttt{rt/arm\_sdk}, \texttt{rt/dex3/left/cmd}, and \texttt{rt/dex3/right/cmd}. We use position control mode with $k_p = 60.0$, $k_d = 1.5$, and a velocity limit of 2.0 rad/s.

\section{EXPERIMENTS}
A video attachment for the experiments is available at 

\url{https://linqi-ye.github.io/video/g1grasp.mp4}

\subsection{Experimental Setup}

All experiments use a computing platform equipped with an NVIDIA GeForce RTX~4060 Laptop GPU for perception and inference. The camera is a RealSense~D435i ($640{\times}480$ at 30 fps), and the robot is a physical Unitree~G1 humanoid with its Dex3 dexterous hand. The target object is a cylindrical drink bottle with a height of 22 cm and a diameter of 6 cm. The workspace is a $20{\times}20$ cm area on a desk located 70 cm in front of the robot.

\subsection{Perception Accuracy}

Table~\ref{tab:yolo} summarizes the detection performance of YOLOv8n on the held-out validation set (30 images). The model achieves mAP@0.5 = 0.995, mAP@0.5:0.95 = 0.858, precision 1.000, and recall 0.997. On an Intel Core~i7-14650HX CPU, single-frame inference takes 30.8 ms (approximately 32 fps), meeting the throughput requirement of the 30 fps RGB-D camera. The model contains only 3.00 M parameters and 8.1 GFLOPs; training for 50 epochs on the same hardware takes only 11 minutes, confirming the design goal of a rapid deployment pipeline.

\begin{figure}[htbp]
\centering
\includegraphics[width=0.95\columnwidth]{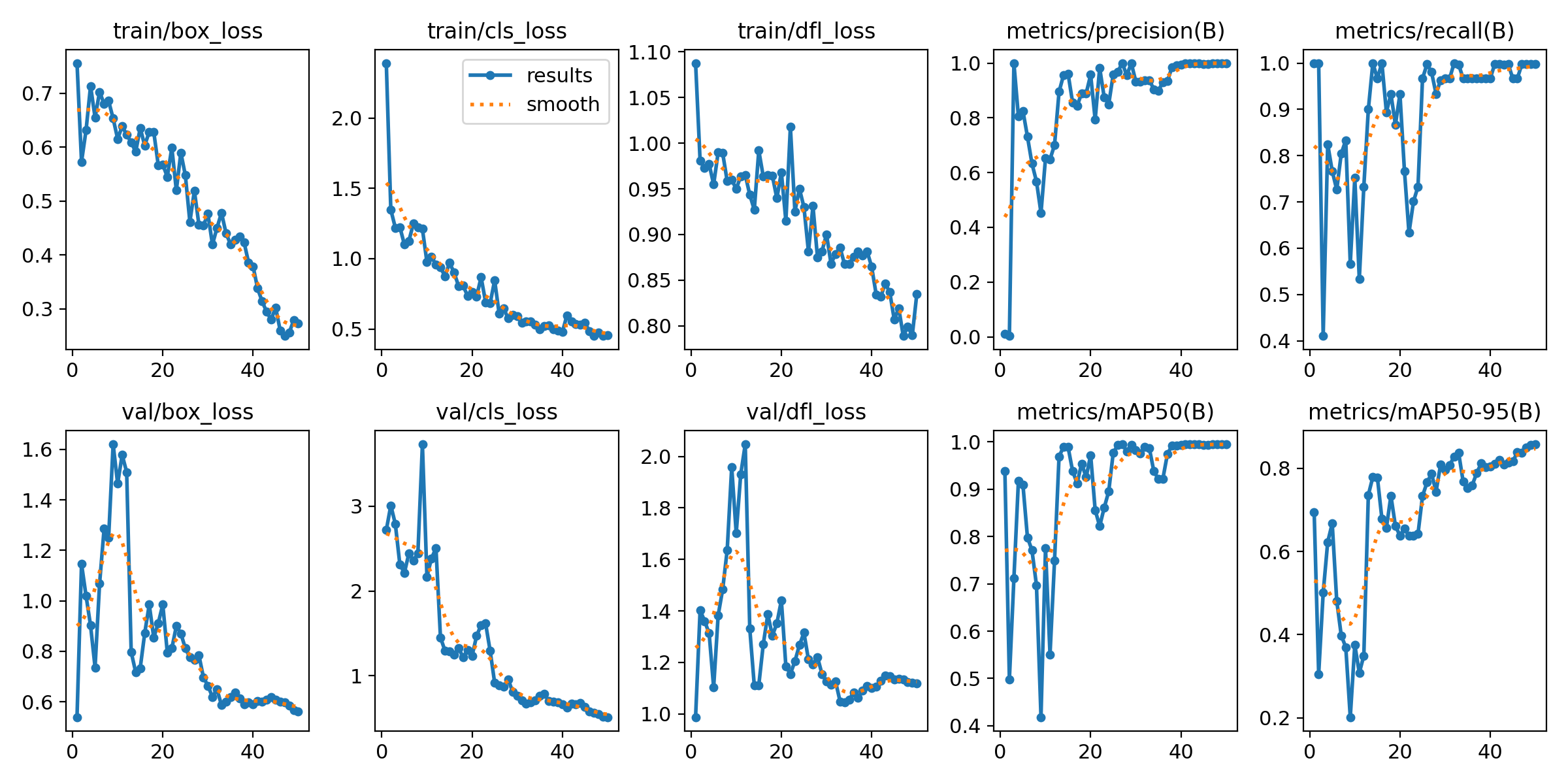}
\caption{YOLOv8n training losses and mAP curves.}
\label{fig:training}
\end{figure}

\begin{figure}[htbp]
\centering
\includegraphics[width=0.95\columnwidth]{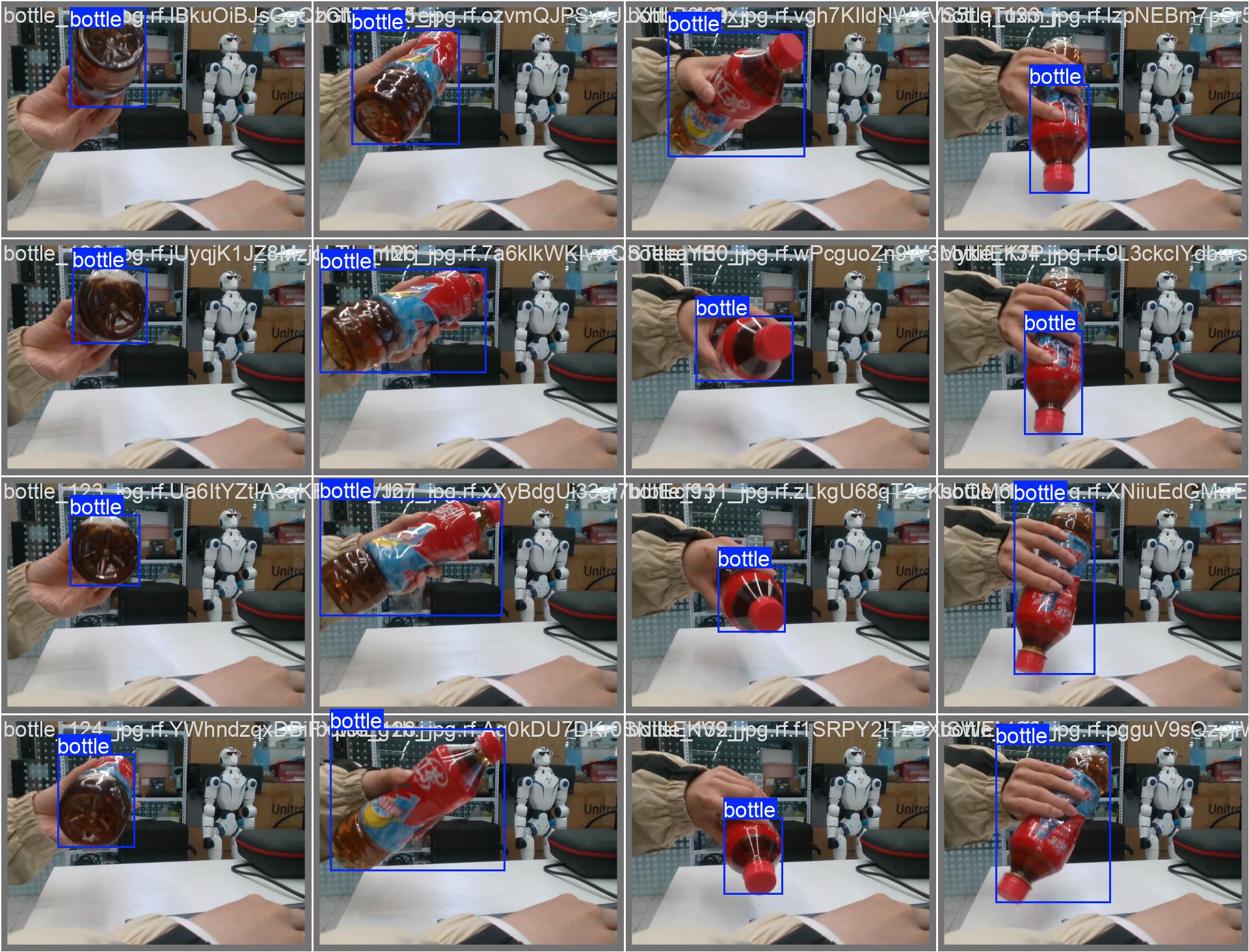}
\caption{YOLOv8n predictions on the validation set.}
\label{fig:preds}
\end{figure}

\begin{figure}[htbp]
\centering
\includegraphics[width=0.99\columnwidth]{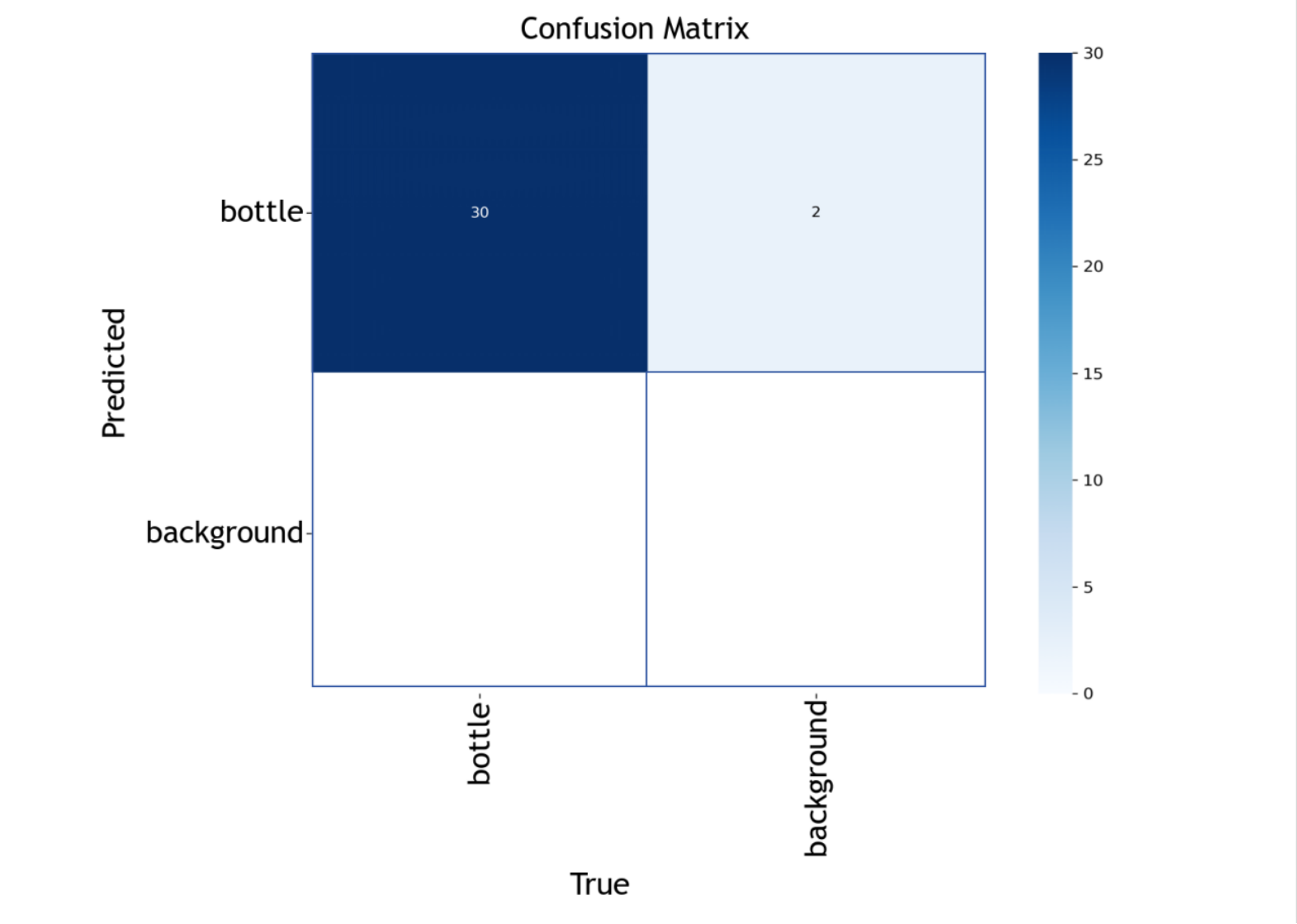}
\caption{YOLOv8n confusion matrix on the validation set.}
\label{fig:confusion}
\end{figure}

Fig.~\ref{fig:training} shows the training curves of YOLOv8n: the losses decrease steadily over iterations, and mAP@0.5 and mAP@0.5:0.95 converge to 0.995 and 0.858, respectively, indicating proper convergence and task-sufficient detection performance. Fig.~\ref{fig:preds} visualizes detection results on the validation set: the model correctly localizes the bottle under varying viewpoints and poses, with no obvious missed or false detections, providing reliable input for subsequent pose tracking. The confusion matrix in Fig.~\ref{fig:confusion} indicates that the model separates the target from the background well, with only a few background false positives and no missed targets, demonstrating good robustness to distractors.

\begin{table}[htbp]
\centering
\caption{Detection Performance of YOLOv8n on the Validation Set}
\label{tab:yolo}
\begin{tabular}{lc}
\toprule
\textbf{Metric} & \textbf{Value} \\
\midrule
Precision & 1.000 \\
Recall & 0.997 \\
mAP@0.5 & 0.995 \\
mAP@0.5:0.95 & 0.858 \\
Single-frame inference time & 30.8 ms \\
Number of parameters & 3.00 M \\
Computation & 8.1 GFLOPs \\
Training time (50 epochs, CPU) & approximately 11 min \\
\bottomrule
\end{tabular}
\end{table}

\subsection{Pose Tracking Performance}

We evaluate pose-tracking performance across three scenarios, with the key metrics summarized in Table~\ref{tab:pose}. In the static scenario, the system achieves sub-millimeter localization precision ($\sigma_{xyz} = 1.05$ mm), with rotation error concentrated along the object's axis of symmetry (the Y-axis accounts for 69.5\%); this geometric ambiguity has no substantive impact on grasping. In the dynamic (handheld) scenario, the system maintains a 100\% tracking success rate with no losses or re-initializations. Under partial occlusion, localization precision drops slightly ($\sigma_{xyz} = 6.40$ mm), but tracking remains stable. Across all scenarios, the system runs at approximately 15--17 fps in debug mode; with visualization disabled, it can match the 30 Hz synchronization requirement of the Unity planner.

\begin{table}[htbp]
\centering
\caption{Pose Tracking Performance Comparison}
\label{tab:pose}
\setlength{\tabcolsep}{4pt}
\small
\begin{tabular}{lccc}
\toprule
\textbf{Metric} & \textbf{Static} & \textbf{Dynamic} & \textbf{Partial} \\
& & \textbf{(handheld)} & \textbf{occlusion} \\
\midrule
Frames                   & 1312   & 1097   & 921    \\
Success rate             & 100.0\% & 100.0\% & 100.0\% \\
$\sigma_{xyz}$ (mm)      & 1.05   & N/A    & 6.40   \\
$\sigma_{\text{rot}}$ ($^\circ$) & 34.14  & 48.70  & 30.55  \\
Y-axis ratio             & 69.5\% & ---    & 66.9\% \\
FPS                      & 15.71  & 16.75  & 17.29  \\
Re-init count            & 0      & 0      & 0      \\
\bottomrule
\end{tabular}
\end{table}

\subsection{End-to-End Grasping on the Physical Robot}

To validate the end-to-end usability of the pipeline, we conducted grasping experiments on the physical Unitree~G1 robot. The workspace is a $20{\times}20$ cm area on a desk 70 cm in front of the robot. Five representative positions were selected (center, front-left, front-right, rear-left, rear-right) to evaluate robustness to target-position variation. At each position, a drink bottle was placed and the complete grasping sequence was executed: the perception system tracks the 6-DoF pose of the object; the Unity IK planner generates a multi-stage grasping trajectory; the C++ bridge issues joint commands through the Unitree SDK; and the Dex3 dexterous hand grasps and lifts the bottle by more than 5 cm.

\begin{figure}[htbp]
\centering
\includegraphics[width=0.95\columnwidth]{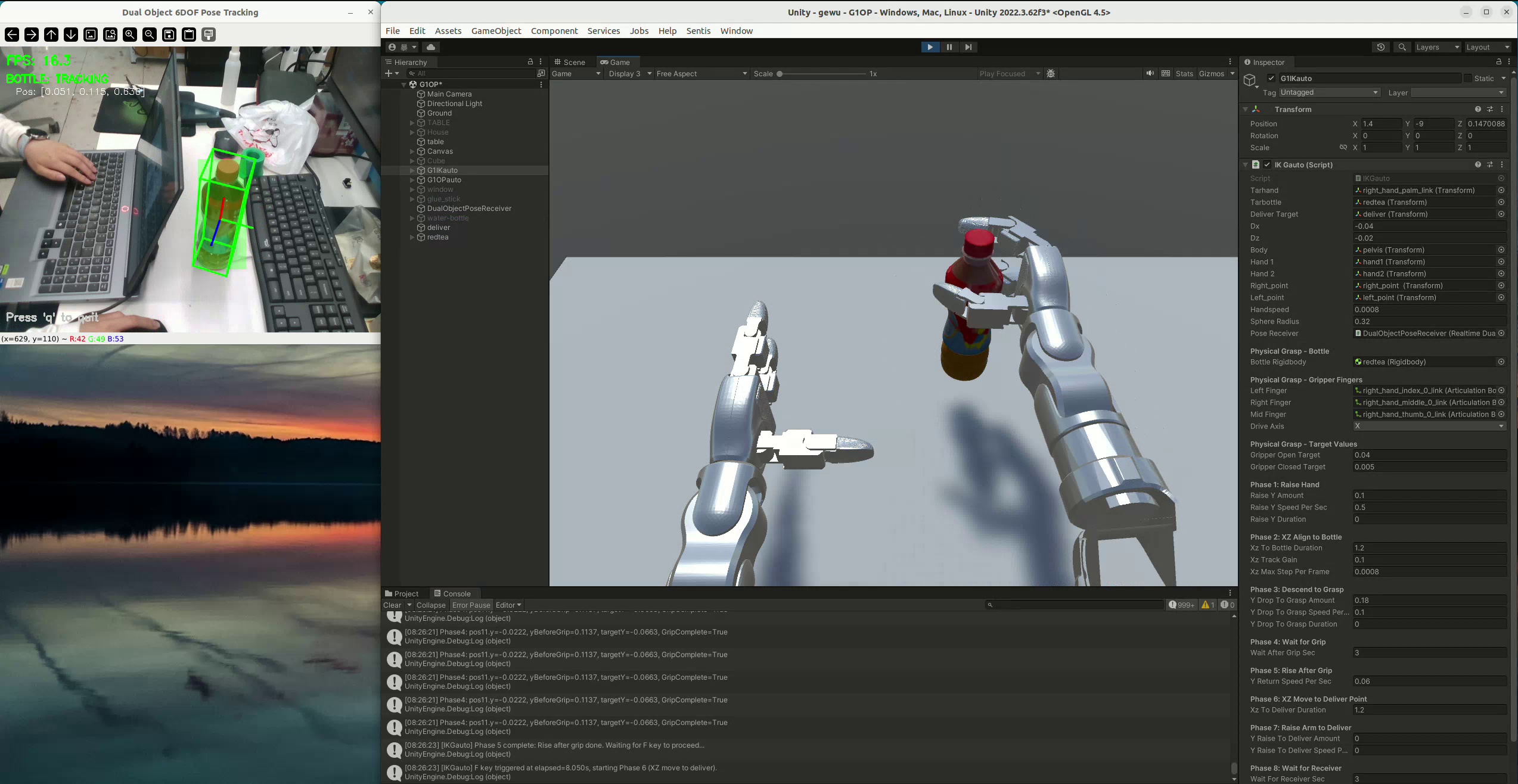}
\caption{Unity simulation of the bottle-grasping task on the G1 URDF.}
\label{fig:sim}
\end{figure}

Fig.~\ref{fig:sim} shows a representative frame from the Unity simulation of the bottle-grasping task. The virtual twin of the tracked object, updated in real time from the perception endpoint, is used by the IK planner to generate the waypoint trajectory; the same normalized joint command stream is what gets transmitted to the physical G1, which allows us to preview and debug the grasping motion in simulation before issuing it to the real hardware.

\begin{figure}[htbp]
\centering
\includegraphics[width=0.95\columnwidth]{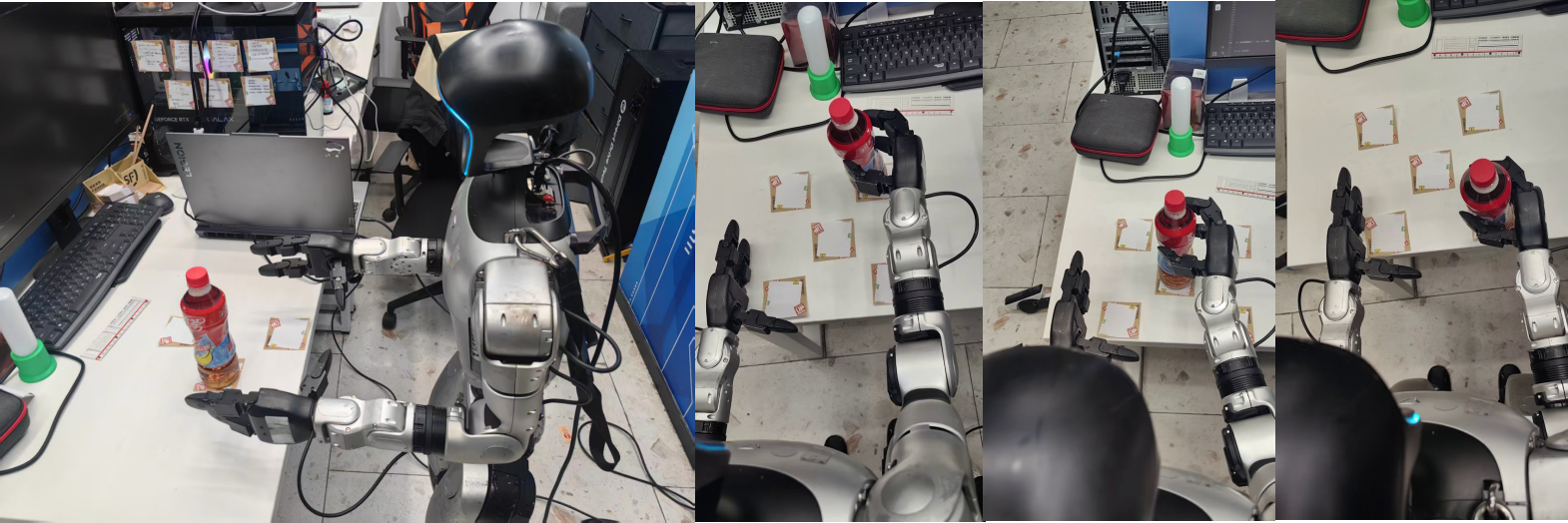}
\caption{End-to-end grasping demonstrations of the Unitree~G1 at five positions within the workspace.}
\label{fig:grasp}
\end{figure}

Fig.~\ref{fig:grasp} shows snippets of successful grasps at the five positions. The robot reliably completes end-to-end grasping at all locations, validating the correct operation of the closed-loop system. The results further confirm the feasibility of the pipeline: first, even in the presence of rotation ambiguity around the symmetry axis, the system still stably grasps cylindrical objects, validating the engineering decisions made at the planning layer; second, meshes produced by monocular SAM~3D reconstruction can be used directly as FoundationPose templates, removing the need for high-precision CAD models obtained via laser scanning and thereby lowering the deployment barrier.

\subsection{Deployment Time Savings}

From taking the first photograph to completing the first successful grasp, the total onboarding time for a new object is about 30 minutes. The breakdown is as follows: data collection approximately 5 min, automatic annotation approximately 5 min, YOLO training approximately 10 min, SAM~3D reconstruction approximately 5 min, and Unity scene configuration approximately 5 min. In contrast, in a traditional pipeline the laser-scanning stage alone typically requires several hours and manual post-processing, and manual annotation of 200--300 images takes several hours.

\subsection{Extended Task: Automobile Window Glue Application}

\begin{figure}[htbp]
\centering
\includegraphics[width=0.6\columnwidth]{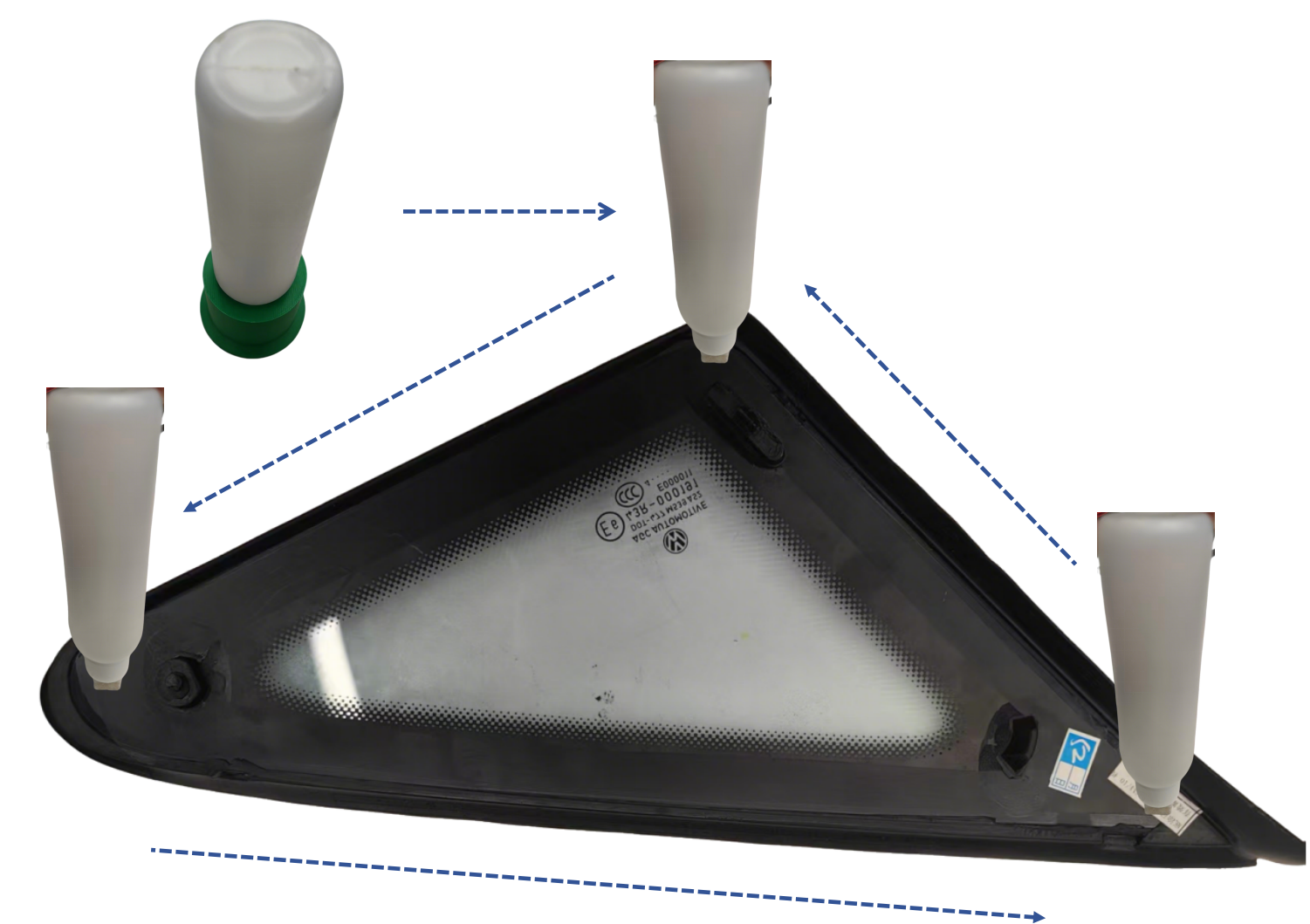}
\caption{ Process of the Gluing Task.}
\label{fig:schematic diagram}
\end{figure}

\begin{figure}[htbp]
\centering
\includegraphics[width=0.95\columnwidth]{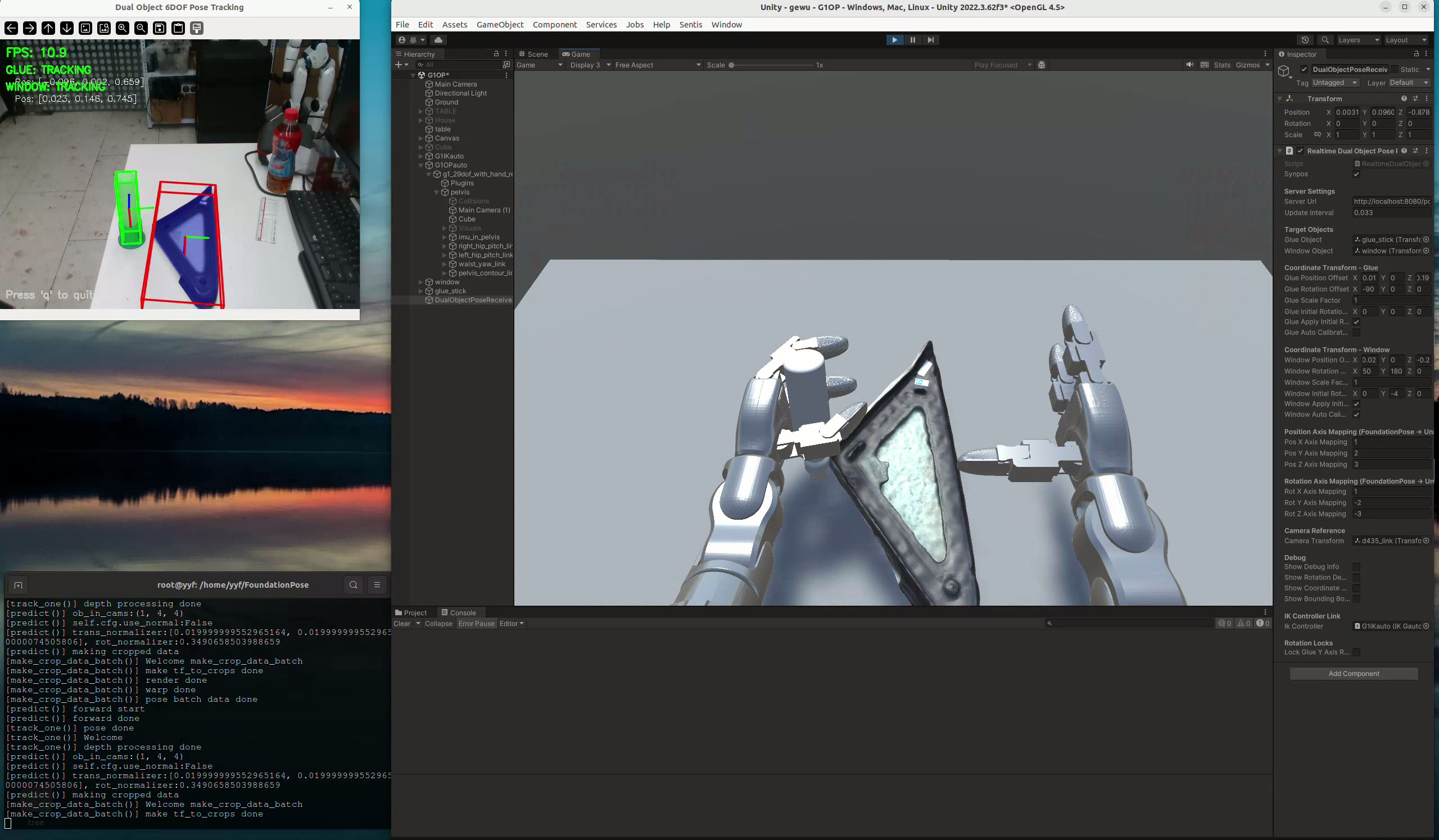}
\caption{Unity simulation of the glue-application task on the G1 URDF.}
\label{fig:glue_sim}
\end{figure}

\begin{figure}[htbp]
\centering
\includegraphics[width=0.95\columnwidth]{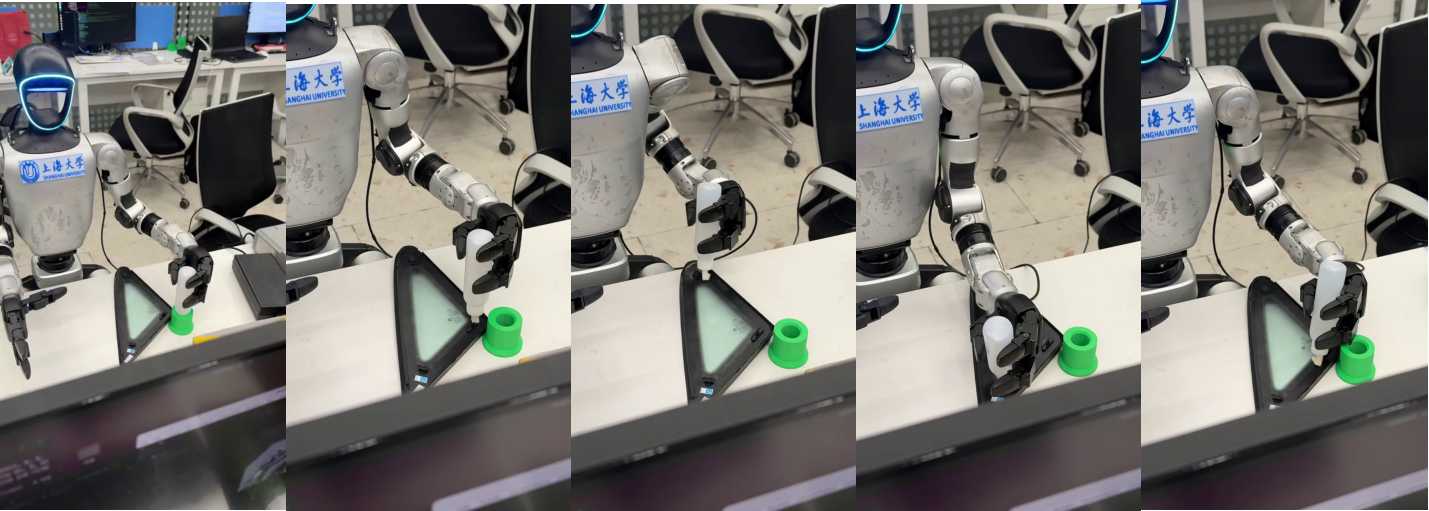}
\caption{A consecutive frame sequence of the window glue-application extension task.}
\label{fig:glue}
\end{figure}

To test the generality of the pipeline, we reuse the same architecture on a structurally very different task: applying glue along the inner edge of a car's rear quarter window using a glue stick, as shown in Fig.~\ref{fig:schematic diagram}. The only modifications are: (i) replacing the object model; (ii) retraining YOLO on a glue-stick and window dataset; and (iii) writing a new trajectory template in Unity. The trajectory is first validated in a Unity-based simulation environment, as illustrated in Fig.~\ref{fig:glue_sim}, and then executed on the physical robot, with a consecutive frame sequence of the real-world glue-application trajectory shown in Fig.~\ref{fig:glue}. The result demonstrates that our pipeline can be transferred to tasks beyond grasping without any algorithmic modification.

\section{DISCUSSION AND LIMITATIONS}

Although our pipeline demonstrates notable advantages in onboarding time for new objects and in end-to-end reliability, several limitations remain and point to directions for future work.

\textbf{SAM~3D handling of reflective and transparent objects.} In early experiments, we observed that SAM~3D produces less stable reconstructions for glossy, highly reflective, or semi-transparent objects (e.g., uncoated plastic bottles, glass bottles), where the generated mesh occasionally exhibits local surface collapse or thickness distortion. For such objects, the current fallback is to supply a small amount of manually authored CAD models (as was done for the glue stick in the window-gluing task), which remains a not-yet-fully-automated step.

\textbf{Robustness under full occlusion.} FoundationPose performs robustly under partial occlusion, but when the object is fully occluded for more than the 3-second threshold set by the state machine, the system transitions to the \textsc{Lost} state. Recovery then relies on YOLO re-detecting the object; if the object's pose upon re-appearance differs substantially from the training distribution, re-registration may fail. Introducing multi-view prompts or a temporal pose-extrapolation mechanism is a worthwhile direction to explore.

\textbf{Quantification of the sim-to-real gap.} We currently use real-robot success rate as the end-to-end metric, but have not systematically quantified the discrepancy between trajectories planned in the Unity simulation and those actually executed on the physical robot. Future work will incorporate joint-level trajectory comparison to more finely identify the sources of error in sim-to-real transfer.

\section{CONCLUSIONS}

This paper proposes and validates a rapid deployment pipeline for humanoid robots that integrates three categories of foundation models---Roboflow-assisted YOLOv8 automatic annotation, Meta SAM~3D 3D reconstruction, and FoundationPose zero-shot 6-DoF pose tracking---with Unity-based inverse-kinematics planning and Unitree SDK physical execution. The complete onboarding time for a new object, from first photograph to first successful grasp, is compressed to approximately 30 minutes. We further verify the generality of the pipeline through an automobile-window glue-application task. The experiments show that combining foundation models with commodity imaging equipment can substantially lower the deployment barrier for humanoid manipulation tasks, providing a foundation for future practical applications in unstructured environments.

\section*{ACKNOWLEDGMENT}

This work was supported by the Technology Commission of Shanghai Municipality (24511103304).

\end{document}